\documentclass[conference]{IEEEtran}
\IEEEoverridecommandlockouts
\usepackage{cite}
\usepackage{amsmath,amssymb,amsfonts}
\usepackage{mathtools}
\usepackage{algorithm}
\usepackage{algpseudocode}
\usepackage{graphicx}
\usepackage{textcomp}
\usepackage{xcolor}
\def\BibTeX{{\rm B\kern-.05em{\sc i\kern-.025em b}\kern-.08em
    T\kern-.1667em\lower.7ex\hbox{E}\kern-.125emX}}
\begin{document}

\title{Improving the Interpretability of Deep Neural Networks with Knowledge Distillation}

\author{\IEEEauthorblockN{Xuan Liu\IEEEauthorrefmark{1},
Xiaoguang Wang\IEEEauthorrefmark{1}\IEEEauthorrefmark{2},
Stan Matwin\IEEEauthorrefmark{1}\IEEEauthorrefmark{3}}
\IEEEauthorblockA{\IEEEauthorrefmark{1}Institute
for Big Data Analytics\\
Faculty of Computer Science, Dalhousie University, Halifax, NS, Canada\\
Email: xuan.liu@dal.ca}
\IEEEauthorblockA{\IEEEauthorrefmark{2}Alibaba Group, Hangzhou, China\\
Email: xiaoguang.wxg@alibaba-inc.com }
\IEEEauthorblockA{\IEEEauthorrefmark{3}Institute of Computer Science\\
Polish Academy of Sciences, Warsaw, Poland\\
Email: stan@cs.dal.ca}}

\maketitle

\begin{abstract}
Deep Neural Networks have achieved huge success at a wide spectrum of applications from language modeling, computer vision to speech recognition. However, nowadays, good performance alone is not sufficient to satisfy the needs of practical deployment where interpretability is demanded for cases involving ethics and mission critical applications. The complex models of Deep Neural Networks make it hard to understand and reason the predictions, which hinders its further progress. To tackle this problem, we apply the Knowledge Distillation technique to distill Deep Neural Networks into decision trees in order to attain good performance and interpretability simultaneously. We formulate the problem at hand as a multi-output regression problem and the experiments demonstrate that the student model achieves significantly better accuracy performance (about 1\% to 5\%) than vanilla decision trees at the same level of tree depth. The experiments are implemented on the TensorFlow platform to make it scalable to big datasets. To the best of our knowledge, we are the first to distill Deep Neural Networks into vanilla decision trees on multi-class datasets.      
\end{abstract}


\section{Introduction}
Despite Deep Neural Networks' (DNN) superior discrimination power in many fields, the logics of each hidden feature representations before the output layer still remain to be black-box. Understanding why a specific prediction is made is of utmost importance for the end-users to trust and adopt the model, and for the system designers to refine the model by performing feature engineering and parameter tuning. This is especially true for high stakes domains such as clinical decision support, disaster response and recidivism prediction. 

For instance, decision trees are preferred over DNN in the health care domain for disease diagnosis due to their ease of interpretation \cite{b1} \cite{b2} \cite{b3}. However, decision trees overfit easily and performs bad on large heterogeneous electronic health records (EHR) datasets \cite{b4}. It is therefore desirable to develop models to find a spot where both interpretability and performance could be ultimately optimized. In recent years, we found resurgent interest in designing interpretable machine learning models. It should be noted that interpretability or transparency of a model is still not clearly defined in the literature \cite{b5} \cite{b6}. 

An intuitive and natural way to interpret neural networks is through visualization. There is a number of works done in this area \cite{b7}. In \cite{b8} two tools are introduced: one plots the activations produced on each layer of a trained neural network; the other visualizes the learned features computed by each neurons at each layer of a neural network. A review of visualization methods for interpreting deep convolutional neural nets is provided in \cite{b9}. However, recent research \cite{b10} shows that it is space, not the individual units, that contains the semantic information in the higher layers of neural networks , which means that the common approach: activation maximization \cite{b7} \cite{b11} \cite{b12} \cite{b13} \cite{b14} applied previously for interpretation has flaws. A related suggestion was given in \cite{b15} to abandon the idea of inspecting individual hidden units. Thus alternative solutions for interpretation are required.

Network diagnosis \cite{b16} is another approach. Earlier research in this area focuses on designing inherently interpretable models such as decision lists \cite{b17}, decision sets \cite{b18}, additive models\cite{b19}, sparse linear models \cite{b20}, etc. However, this approach presents a severe constraint on the selection of algorithms. Besides, although human can comprehend these models, they fail to model more complex problems with good accuracy performance. 

In this paper, we apply the most recent model-agnostic approach \cite{b21} which performs post-hoc explanations on the trained models. Past research focus on either global interpretations \cite{b22} \cite{b23} or local explanations \cite{b24}\cite{b25}\cite{b26}. We concentrate on the global interpretations. In this paper, we adopt knowledge distillation to improve the global interpretation results.

Knowledge distillation refers to the process of transferring the dark knowledge learned by a teacher model (usually sophisticated and large) to a student model (usually shallow and small). Dark Knowledge \cite{b27} \cite{b28} is the salient information hidden in the ``soft targets'': predicted probabilities for all classes, which are more informative than the ``hard targets'': predicted classes. Maybe the pioneer work to distill the knowledge from a neural network into another algorithm is by Craven and Shavlik \cite{b29} who used a symbolic algorithm: the decision tree \cite{b30} to approximate the functions learned by a neural network with one hidden layer using hard targets. 

Knowledge distillation originates from model compression \cite{b31}. In \cite{b31}, the teacher model was built using the ensemble selection algorithm \cite{b32}, which was then used to label unseen unlabeled data: the training data for the student model (also called the transfer data). This approach uses the hard targets produced by the teacher model. A followed work \cite{b33}  distills deep nets into shallow feed-forward nets adopting the method of ``matching logits'' (scores before the softmax activations), which would avoid the information loss when passing through logits to the probability space. Then the concept of ``knowledge distillation'' was officially introduced in \cite{b27}. It is a more general solution to transfer knowledge from a cumbersome model to a compact model. They try to find an optimal temperature (which they inset into the term of the softmax layer) by raising the temperature of the final softmax layer of the teacher model until a suitable set of soft targets are generated. Then they apply the same temperature to the student model. They also proved that ``matching logits'' was actually a special case of their distillation approach.

Afterwards, a number of works followed such as \cite{b34}\cite{b35}\cite{b36}, just to name a few. Most of these works concentrate on distilling complex and deep neural nets into simple and shallow neural nets. And are mainly applied for scenarios like edge computing hardware and on-the-fly training where there are  memory, resource, power, time and space constraints, without significant loss in performance. 

In our work, we employ knowledge distillation for another purpose: interpretation. We resolve the tension between interpretability and accuracy performance by distilling deep neural nets into vanilla decision trees. This is a work in progress and as the first step of our attempts we apply the matching logits approach in \cite{b33}. The main obstacle to execute this plan is that for pure classification tasks there exist no logits in decision trees as in neural nets which could be used in the loss function. We address this issue by reformulating it into a multi-output regression problem \cite{b37} and achieved significant accuracy improvements (about 1\% to 5\%) on the experiments. Hence, the success of our approach opens a door for turning those inherently interpretable algorithms (which are highly interpretable, but worse in accuracy performance) into models attaining both accuracy and interpretability simultaneously. 

\section{Related Work}
Perhaps the most related work is the model in \cite{b15} which uses a type of soft decision tree to mimic the input-output functions of a trained DNN. This soft decision tree produces hierarchical decisions, which is more easier to interpret than DNN that relies on hierarchical features. It is modeled based on hierarchical mixture of experts and trained with gradient descent. The way they design the soft decision tree is quite similar to \cite{b38}. Knowledge distillation was then used to improve the soft decision tree's accuracy. The difference between their approach and ours is that we use vanilla decision tree as the student model while their student model is the soft decision tree which has similar architecture with neural networks and could be more easily adapted to the original knowledge distillation framework.

In the health care domain, two pipelines \cite{b41}\cite{b4} are proposed to distill the knowledge from a DNN to Gradient Boosting Trees (GBTs) \cite{b39}\cite{b40}. One of them extracts the logits from a learned DNN and uses the logits and the true labels of the original training data to train a logistic regression algorithm to obtain the soft prediction scores. The next step is to train GBTs with the original training data's features and the soft predictions. The second pipeline directly applies the soft prediction scores of the trained DNN on the original training data as targets for training a mimic model with GBTs. However, GBTs lack transparency as they rely on post-hoc determinations: partial dependence \cite{b39}, which would result in bias in this process \cite{b49}.  The differences between their approach and ours are apparent. The strategy we applied when training the mimic model is matching logits, not the soft targets. Also, our student model is decision tree. 

Another approach that distills neural networks into GBTs is in \cite{b42}. They tried two student models: tree-based generalized additive models (GA2Ms) \cite{b19}\cite{b43}\cite{b44} and GBTs. The teacher model they adopted is multilayer perceptrons. For the student model's training process, they applied the method of matching logits instead of soft targets in \cite{b41}\cite{b4}. They investigated both classification and regression problems. However, their model is limited to the binary class problems and their results are not conclusive yet and not published. Compared to their method, our teacher model is DNN and the student model is decision tree. We aims at multi-class classification problems. 

Instead of doing post-hoc interpretations, this work \cite{b45} focuses on finding more interpretable neural networks during the training process. They created a new model complexity penalty function: tree regularization to favor models whose decision boundaries could be well approximated by small decision trees. They measure human simulatability as the average decision path length and make the decision tree loss differentiable by adopting the technique of derivative-free optimization techniques \cite{b46}. Their experiments show that using tree regularization could achieve high accuracy at low complexity. Our method belongs to the post-hoc interpretations, which is different from what they proposed.

A most recent work \cite{b47} combines knowledge distillation and dimension reduction to visualize the results of deep classifiers. They pointed out that the method: t-distributed stochastic neighbor embedding (t-SNE) \cite{b48} commonly used for visualizing the activations of hidden layers was problematic. They propose to visualize the data points that are assigned similar probability vectors to give practitioners a sense of how the decisions are made on test cases. They train a simpler and more interpretable classifier using the soft targets generated by a deep classifier. The student model they applied is Naive Bayes. 

\section{Methodology}
Rather than common approaches that distill DNN into shallow neural networks, we investigate the distillation into non-neural nets. And the deep models we focus on is Convolutional Neural Networks (CNN). We first introduce some background information about CNN, decision trees and knowledge distillation and then describe our own methodology in details. 
\subsection{Convolutional Neural Networks}
The architecture of CNN is similar to the LGN--V1--V2--V4--IT  hierarchy in the visual cortex ventral path-way \cite{b50}. It is designed to process data that has a known, grid-like topology, e.g. image data that has a 2D grid of pixels. Its typical framework is a stack of convolutional-pooling layers followed by fully connected layers. Also, the results of the convolutional layer has to pass through a nonlinear activation function. A commonly used one is the rectified linear unit ReLU \cite{b51}. The convolutional and pooling layers originates from the concepts of simple cells and complex cells in visual neuroscience \cite{b52}.

\begin{itemize}
\item \textbf{Convolutional layer}. The feature maps of the convolutional layer are generated by performing discrete convolutions between a series of weights and the results of the previous layer. These weights are named as filter banks \cite{b53} or kernels \cite{b54}. For 2-D grayscale inputs, the value of a specific unit $x_{ij}^l$ of feature map $k$ in the first convolutional layer $l$ with kernel size $m\times m$ is calculated as
\begin{equation}
 x_{ij}^l=\sum_{a=0}^{m-1}\sum_{c=0}^{m-1}w_{ab}o_{(i+a)(j+c)}^{l-1}+b_k\label{eq1}
\end{equation}
The first term in (1) is the convolution operations and the second term is the bias for this feature map. If there are multiple channels of the input image, the first term will be summed over all these channels to produce one unit in the corresponding feature map. Within a feature map, all units share the same filter bank and bias. The convolutional operation is accomplished after the filter bank slide across the width and height of the input image.
\item \textbf{ReLU}. Following the convolutional layer is a non-saturating nonlinearity function: ReLU. For a specific unit $x_{ij}^l$, its values after passing through this function is 
\begin{equation}
o_{ij}^l=max(0,x_{ij}^l)\label{eq2}
\end{equation}
It was reported \cite{b55} that for gradient descent training, using ReLU could speed up the training time several times faster than saturating nonlinearities such as $tanh$. 
\item \textbf{Pooling layer}. It works as a down-sampling tool and merges semantically similar units into one. At this layer, the value of the output unit is a summary statistic of the nearby outputs of the previous layer. Usually, the \textit{max pooling} function \cite{b56} is applied here. It outputs the maximum value within a rectangular area. For instance, if this rectangular area has size $k\times\ k$ and one feature map of the previous layer has size $N\times\ N$. The resulting feature map will have size $\frac{N}{k} \times \frac{N}{k}$. Pooling layer reduces the dimensions of representations, hence, helps speed up the training process. In addition, it helps to make the feature maps invariant to small shifts and distortions of the inputs. 
\end{itemize}
In the training process, CNN performs backward propagation similar to  the regular fully connected networks so that all the weights could be updated.
\subsection{CART for Regression} 
There are several versions of the decision tree algorithm. The earliest version: Iterative Dichotomiser 3 (ID3)\cite{b57} was proposed by Quinlan in 1986. It uses information gain as its attribute selection measure and requires features to be categorical. C4.5\cite{b58} is a successor of ID3 by Quinlan and the restrictions of ID3 on features are removed. Classification and Regression Trees (CART) \cite{b59} was introduced in 1984 by Breiman \textit{et al}. Although CART and C4.5 were invented by different authors, they follow similar ideas for training decision trees. Owing to the reason that CART supports numerical target values  (regression) and the key to our methodology is to solve a multi-output regression problem, we introduce briefly here the algorithm of CART.
CART applies a greedy approach which constructs the decision tree in a top-down recursive divide-and-conquer manner. As our experiments applies CART for regression, the descriptions focus on regression tasks.

This algorithm partitions the feature space and groups instances with the same labels together. Initially, it constructs a root node with all training samples {S} with features as $x_i\in R^n$ for $i=1...l$ and labels as $y_i\in R^l$  and split the node into two child nodes recursively. The splitting criterion is: $C=(a,t_n)$, where $a$ is the attribute to split on and $t_n$ is the threshold at node n. This criterion partitions $S$ into
\begin{equation}
S_{left}{(C)}=(x,y)|x_a\leqslant t_n\label{eq3}
\end{equation}
\begin{equation}
S_{right}{(C)}=S\setminus S_{left}{(C)}\label{eq4}
\end{equation}
The impurity at node $n$ is calculated with an impurity function $I$. For our regression task, we applied the Mean Squared Error method to calculate the impurity. Hence, $I$ is calculated as
\begin{equation}
y_n^{\prime}=\frac{1}{M_n}\sum_{i\in M_n} y_i\label{eq5}
\end{equation}
\begin{equation}
I(X_n)=\frac{1}{M_n}\sum_{i\in M_n} (y_i-y_n^{\prime})^2\label{eq6}
\end{equation}
$M_n$ is the number of instances in the corresponding child node. Hence, based on $I$, the impurity for both nodes can be expressed as
\begin{equation}
f(S,C)=\frac{M_{left}}{M_n}I(S_{left}(C))+\frac{M_{right}}{M_n}I(S_{right}(C))\label{eq7}
\end{equation}
Then the parameters in {C} could be optimized by minimizing $f(S,C)$
\begin{equation}
C^{\ast}=argmin_Cf(S,C)\label{eq8}
\end{equation}
Thus, the optimal attribute and the splitting threshold are found. Then the algorithm recursively splits $S_{left}{(C)}$ and $S_{right}{(C)}$ until the maximum depth specified by the user is reached, a node becomes pure, $M_n<min_{samples}$ or  $M_n=1$.

\subsection{Matching Logits} 
\begin{figure}[htbp]
\centerline{\includegraphics[scale=0.45]{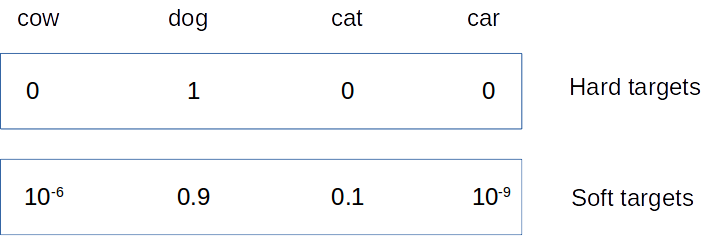}}
\caption{Examples of hard and soft targets.}
\label{fig_1}
\end{figure}

Knowledge distillation transfers the generalization ability of a complex teacher model to a simple student model. Using the teacher model's soft targets for distillation could produce much better outcome than hard targets. Fig.~\ref{fig_1} shows an example of hard and soft targets. Hard targets just contain the information for the predict label while soft targets reveal all the predicted probabilities for all the classes. Many previous works \cite{b22} \cite{b31} \cite{b60} just adopt the hard targets (the predicted labels of the teacher model) for distillation, where soft targets could as a matter of fact boost the results significantly. 

When we exam closely into Fig.~\ref{fig_1}, we notice that the probabilities for ``cow'' and ``car'' are much smaller than those of ``dog'' and ``cat''. When training student models applying the cross-entropy cost function, these much smaller probabilities would vanish to zero. Take CNN for example, the last hidden layer $l$ before the softmax layer is a fully connected layer with logits $z$ as the output
\begin{equation}
z_i=\sum_jW_{ij}x_j^{l-1}+b_j\label{eq9}
\end{equation}
Here $z_i$ is the logit for one of the classes: $i$. $j$ is the number of hidden nodes for layer $l-1$. $W$ and $b$ are weights and bias respectively. The softmax layer calculates the output probabilities for each class as
\begin{equation}
q_i=\frac{e^{z_i}}{\sum_je^{z_j}}\label{eq10}
\end{equation}
The cross-entropy function is then applied to calculate the loss of the model
\begin{equation}
H_{p}(q)=-\sum_ip_ilog(q_i)\label{eq11}
\end{equation}
Hence, to avoid the loss of information, it is desirable to use logits $z$ instead of the predicted probabilities $q$. This method is called ``matching logits'' and the pioneer work was done in \cite{b33}. Hinton \textit{et al}. \cite{b27} extended their work to a more general case by inserting a temperature term $T$ into \eqref{eq10}
\begin{equation}
q_i=\frac{e^{\frac{z_i}{T}}}{\sum_je^{\frac{z_j}{T}}}\label{eq12}
\end{equation}
and they demonstrated mathematically that in the high temperature limit and when the logits were zero-meaned separately for each training instance of the student model, matching logits was a special case of using the soft targets for distillation. They proved it by performing gradient descent on the cross-entropy function
\begin{equation}
\frac{\partial H}{\partial z_i}\approx \frac{1}{NT^2}(z_i-v_i)\label{eq13}
\end{equation}

Here $v_i$ is the logit of the teacher model for training instance $i$, $N$ is the number of instances of the training data for the student model. For more elaborated derivations, please refer to \cite{b27}.

\subsection{Distilling CNN into Decision Trees} 
\begin{figure}[htbp]
\centerline{\includegraphics[scale=0.45]{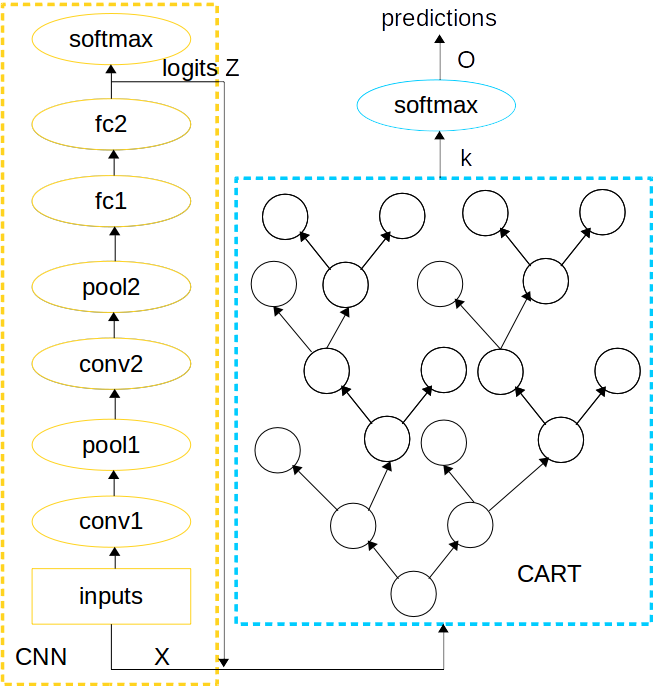}}
\caption{Framework of our method.}
\label{fig_2}
\end{figure}

In this work, as the first step of our attempts, we employ the matching logits method when distilling CNN into vanilla decision trees. Fig.~\ref{fig_2} illustrates the framework of our method. In this figure, the architecture of the CNN is the one used to train the MNIST data as in \cite{b61}. It comprises of two convolutional layers and two pooling layers followed by two fully connected layers: fc1 and fc2. After this deep CNN is trained, we feed the feature part $X$ of the original training data to the trained model to obtain the corresponding logits $Z$. Then we train CART with $X$ and $Z$ which is treated as the targets. 

However, here arise some problems for deployment. First, for classification tasks, the targets are limited to categorical, not numerical and continuous values. For this, we can resolve by treating it as a regression problem. Second, even for regression tasks, most algorithms only support single-output regressions. For multi-class datasets, this is actually a multi-output regression problem \cite{b37}. And we apply the algorithm adaptation method, where we use decision trees to directly handle multi-output data sets simultaneously. This is anticipated to produce much better results than the problem transformation method which transforms the multi-output regression problem into independent single-output problems and are then solved by single-output regression algorithms. This is due to the fact that problem transformation methods don't consider the dependencies among the targets. 

So the key novelty of this paper is that we treat the problem at hand as a multi-output regression problem first and then try to translate the regression results to achieve the goal of classification. Hence, the regression data for CART should have features as $X$ with  $x_i\in R^n$ for $i=1...l$ and labels as $Z$ with $z_i\in R^l$. And the impurity function $I$ in CART is calculated as
\begin{equation}
z_n^{\prime}=\frac{1}{M_n}\sum_{i\in M_n} z_i\label{eq14}
\end{equation}
\begin{equation}
I(X_n)=\frac{1}{M_n}\sum_{i\in M_n} (z_i-z_n^{\prime})^2\label{eq15}
\end{equation}
Once CART is trained, in order to obtain the final prediction results on test cases we need to add a softmax layer over the test results of CART to turn numerical test results into categorical ones. Assuming the test results on CART is $k$, the final output probabilities for class $i$ therefore is
\begin{equation}
O_i=\frac{e^{k_i}}{\sum_je^{k_j}}\label{eq16}
\end{equation}

\section{Experiments}
We performed the experiments on two datasets to demonstrate the effectiveness of our distillation approach. All teacher models are implemented on the TensorFlow \cite{b62} platform to make them scalable to big datasets.
\subsection{Datasets}
The two datasets we selected are the MNIST dataset \cite{b63} and the Connect-4 dataset from the UCI repository \cite{b64}. MNIST is a famous benchmark dataset for deep learning. It contains the pixel values of handwritten digits from 0 to 9. Each instance stands for a $28\times 28$ grayscale image and contains 784 features when flattened into a one dimensional space. The Connect-4 dataset stores the information about the two players' positions for the the game of connect-4. It has a seven-column, six-row vertically suspended grid. There are two players and each spot on the grid represents whether it has been taken by the first player, or the second player or left blank. The classes are the outcome for the first player. Details of these datasets could be found in Table~\ref{tab1}.

\begin{table}[htbp]
\caption{Datasets}
\begin{center}
\begin{tabular}{|c|c|c|c|c|}
\hline
\textbf{}&\multicolumn{4}{|c|}{\textbf{Dataset Details}} \\
\cline{2-5} 
\textbf{} & \textbf{\textit{\#Features}}& \textbf{\textit{\#Train}}& \textbf{\textit{\#Test}} & \textbf{\textit{Labels}}\\
\hline
MNIST & 784 & 55,000 & 10,000 & 0-9\\
\hline
Connect-4 & 42 & 57,557 & 10,000 & {win, loss, draw} \\
\hline
\end{tabular}
\label{tab1}
\end{center}
\end{table}

\subsection{Experimental Setup}
The deep learning model we applied to train the MNIST dataset is a deep CNN which has an architecture of two convolutional layers followed by two fully connected layers. The parameter settings for this network is depicted in Table~\ref{tab2}. The first convolutional layer uses filters with window size $5\times 5$, $stride=1$ and the `same' padding in TensorFlow. When the stride length is 1, `same' padding generates a feature map with the same size as the input image. This stage produces 32 feature maps each with size $28\times 28$. The followed max pooling layer over $2\times 2$ blocks with $stride=2$ generates 32 feature maps with size $14\times 14$. The parameter settings for the second convolutional layer and pooling layer are the same as the previous one except that this stage generates 64 feature maps. Hence, we have 64 feature maps each with size $7\times 7$. Then we flatten these $7\times 7\times 64$ features into a one dimensional list and then apply a fully connected layer: fc1 with 1024 hidden nodes. Immediately after fc1 is the dropout \cite{b65} layer, where we set the dropout rate as 0.5. The second fully connected layer: fc2 is the output layer with 10 hidden nodes, each representing one of the 0-9 digits. These outputs are also the logits of this model. 

The Connect-4 dataset has a class distribution of win (65.83\%), loss (24.62\%) and draw (9.55\%). We randomly sample 10,000 test instances which satisfy the original class distributions. The algorithm we applied to train the Connect-4 dataset is a multilayer perceptron (MLP) with parameter settings in Table~\ref{tab3}. It has three hidden layers, the first hidden layer with 256 hidden nodes, the second hidden layer with 128 hidden nodes, the third hidden layer also with 128 hidden nodes and the output layer with 3 nodes representing the three outcomes of the connect-4 game. We also apply the dropout rate after each of the hidden layers and the value is set as 0.8. When calculating the training loss, in addition to TensorFlow's own cross entropy function, we also added a L2 penalty (regularization term) parameter as in Python's scikit-learn machine learning tool. This penalty parameter is set as 0.0001 which could help to improve the MLP's performance.

\begin{table}[htbp]
\caption{Parameter settings for MNIST}
\begin{center}
\begin{tabular}{|c|c|c|c|c|c|}
\hline
\textbf{}&\multicolumn{5}{|c|}{\textbf{Network Type: CNN}} \\
\cline{2-6} 
\textbf{} & \textbf{\textit{conv:filter}}& \textbf{\textit{conv:stride}}& \textbf{\textit{pool:block}} & \textbf{\textit{pool:stride}} & \textbf{\textit{fc1}}\\
\hline
MNIST & $5\times 5$ & 1 & $2\times 2$ & 2 & 1024\\
\hline
\end{tabular}
\label{tab2}
\end{center}
\end{table}

\begin{table}[htbp]
\caption{Parameter settings for Connect-4}
\begin{center}
\begin{tabular}{|c|c|c|c|c|c|}
\hline
\textbf{}&\multicolumn{5}{|c|}{\textbf{Network Type: MLP}} \\
\cline{2-6} 
\textbf{} & \textbf{\textit{1st hidden}}& \textbf{\textit{2nd hidden}}& \textbf{\textit{3rd hidden}} & \textbf{\textit{out}} & \textbf{\textit{dropout}}\\
\hline
Connect-4 & 256 & 128 & 128 & 3 & 0.8\\
\hline
\end{tabular}
\label{tab3}
\end{center}
\end{table}

\subsection{Experimental Results}
For decision tree classifications, we apply the modules in the scikit-learn machine learning tool. When we are performing classification tasks applying a decision tree, there are a variety of parameters to tune such as the minimum number of samples per leaf, the strategy used to choose the split at each node (either the best split or the best random split) and so on. We select two parameters that would influence the performance of a decision tree substantially: the maximum depth of the tree and the functions to measure the impurity of a split (either ``gini'' or ``entropy''). The other parameters are left as default values as in scikit-learn. 

For the MNIST dataset, the teacher CNN model achieves an accuracy of 99.25\%. The performance for the student model and the vanilla decision tree classification results are shown in Table~\ref{tab4}. ``Acc\_student'' represents the accuracy of the student decision tree trained using the logits of the teacher CNN model on TensorFlow. 

\begin{table}[htbp]
\caption{Test Accuracy Results for MNIST}
\begin{center}
\begin{tabular}{|c|c|c|c|}
\hline
\textbf{Tree}&\multicolumn{3}{|c|}{\textbf{Methods}} \\
\cline{2-4} 
\textbf{\textit{Depth}}& \textbf{\textit{Acc\_student}}& \textbf{\textit{Acc\_gini}} & \textbf{\textit{Acc\_entropy}}\\
\hline
6 & \textbf{0.7119} & 0.6644 & 0.6849\\
\hline
7 & \textbf{0.7685} & 0.7534 & 0.7228 \\
\hline
8 & \textbf{0.8125} & 0.7914 & 0.8007 \\
\hline
9 & \textbf{0.8512} & 0.8151 & 0.8304 \\
\hline
10 & \textbf{0.8655} & 0.8445 & 0.8450 \\
\hline
\end{tabular}
\label{tab4}
\end{center}
\end{table}

\begin{figure}[htbp]
\centerline{\includegraphics[scale=0.45]{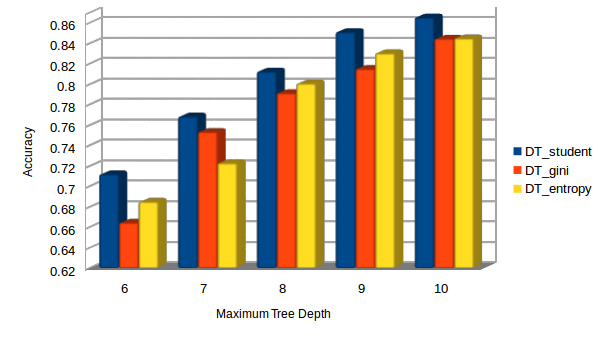}}
\caption{Distillation results for MNIST.}
\label{fig_3}
\end{figure}

\begin{figure}[htbp]
\centerline{\includegraphics[scale=0.45]{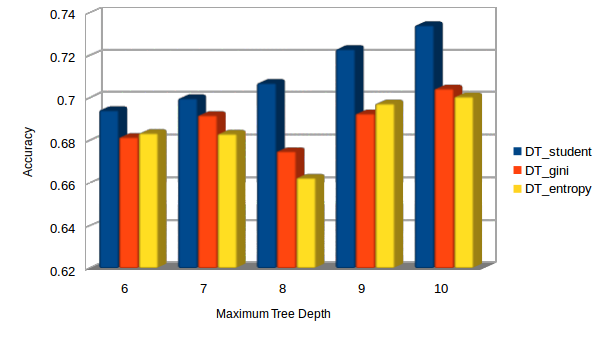}}
\caption{Distillation results for Connect-4.}
\label{fig_4}
\end{figure}

``Acc\_gini'' is the accuracy of the decision tree without distillation when the impurity measure is ``gini'' in scikit-learn when trained utilizing the same training and test data as the CNN model. ``Acc\_entropy'' is the classification accuracy of the decision tree when the impurity measure is ``entropy''. We highlighted the best performance in bold. Under different tree depths, the student model always outperforms the vanilla decision tree. The same conclusion holds true for the Connect-4 dataset in Table~\ref{tab5} where the accuracy for the MLP teacher model is 86.62\%. The reason we limit the tree depth to 10 is that we would like to construct interpretable models and trees over a depth of 10 becomes extremely hard for human cognitions to comprehend. We also illustrate these results in graphs in Fig.~\ref{fig_3} and Fig.~\ref{fig_4} to present the results more intuitively.

\subsection{Discussion}
For the Connect-4 dataset, although we can fine tune the parameters of the teacher model or switch the teacher model to CNN to improve the teacher models' performance, the distillation effect still relies largely on the student model's own generalization ability. For instance, the teacher model for the MNIST dataset already has a very high accuracy of 99.25\%, but the student model's highest accuracy in Table~\ref{tab2} is only 86.55\%. However, from our experiments we found that training a good teacher model indeed helped to boost the distillation results. In our experiments, we notice that distillation helps to improve the accuracy by 1\% to 5\%. Hence, there is still a long way to go for the student model to match the results of the teacher model.

\begin{table}[htbp]
\caption{Test Accuracy Results for Connect-4}
\begin{center}
\begin{tabular}{|c|c|c|c|}
\hline
\textbf{Tree}&\multicolumn{3}{|c|}{\textbf{Methods}} \\
\cline{2-4} 
\textbf{\textit{Depth}}& \textbf{\textit{Acc\_student}}& \textbf{\textit{Acc\_gini}} & \textbf{\textit{Acc\_entropy}}\\
\hline
6 & \textbf{0.6943} & 0.6816 & 0.6835\\
\hline
7 & \textbf{0.6999} & 0.6919 & 0.6832 \\
\hline
8 & \textbf{0.707} & 0.675 & 0.6625 \\
\hline
9 & \textbf{0.723} & 0.6927 & 0.6974 \\
\hline
10 & \textbf{0.7342} & 0.7044 & 0.7006 \\
\hline
\end{tabular}
\label{tab5}
\end{center}
\end{table}

We are also curious about the performances of the student models and the vanilla decision trees when the maximum depth of the tree is not specified. In this situation, for the MNIST dataset, we found that the accuracy for the student model was 88.28\% and the decision tree classification achieved 87.4\% for the criterion of ``gini''. For the Connect-4 dataset, when the teacher model has an accuracy of 83.22\% the student model achieves 79.06\% and vanilla decision tree has 77.57\% when using ``gini'' as impurity measure. We notice that the accuracy improvements are smaller than the cases where the depth of the trees are specified. This is easy to explain as when the tree levels are not set the vanilla decision tree takes much deeper tree levels than the student model to arrive at the current accuracy results. Hence these decision trees are far less interpretable than the student models because the level of tree depth determines the interpretability for decision trees. After all, in our experiments we already proved that under the same tree level, the vanilla decision tree performs worse than the student models.

\section{Conclusion and Future Work}
Based on the fact that inherently interpretable algorithms perform worse than some non-interpretable algorithms such as the deep learning algorithm, this paper presents an approach to improve the accuracy performance of an inherently interpretable algorithm: decision tree. This is achieved by utilizing the dark knowledge hidden in the soft predictions of DNN. We apply the matching logits method which employs the logits of DNN for training student decision tree models. Experiments on two datasets: MNIST and Connect-4 demonstrate the significant improvements on the accuracy of the distilled student model over vanilla decision trees. 

Our work is still in progress and there are several directions for future work. First, as specified in \cite{b37}, there are various methods to solve the multi-output regression problem. The method we adopted is the algorithm adaptation method. It is worthwhile to explore other methods to fully take advantage of the power of knowledge distillation. Second, our approach in this paper makes it possible to improve the performance of all inherently interpretable models and it is therefore rewarding to design new inherently interpretable models that could finally match the performance of non-interpretable models. Last, it should also have merits to add a temperature term into the softmax layer (as introduced in the methodology part) and use both soft targets and the true labels together (as carried out in \cite{b27}) to train the student model.

\section*{Acknowledgment}
The authors would like to thank Xiang Jiang, Zhengping Che, Sarah Tan and Nicholas Frosst\ for helpful discussions.

\end{document}